# RL STaR Platform: Reinforcement Learning for Simulation based Training of Robots

**Tamir Blum[1], Gabin Paillet[1], Mickael Laine[1], Kazuya Yoshida[1]**

[1]*Department of Aerospace Engineering, Tohoku University, Aoba 6-6-01, Aramaki, Aoba-ku, Sendai, Miyagi 980-8579, Japan*

*Emails: tamir@dc.tohoku.ac.jp, paillet.gabin.p5@dc.tohoku.ac.jp, mickael@astro.mech.tohoku.ac.jp yoshida@astro.mech.tohoku.ac.jp*

## ABSTRACT

Reinforcement learning (RL) is a promising field to enhance robotic autonomy and decision making capabilities for space robotics, something which is challenging with traditional techniques due to stochasticity and uncertainty within the environment. RL can be used to enable lunar cave exploration with infrequent human feedback, faster and safer lunar surface locomotion or the coordination and collaboration of multi-robot systems. However, there are many hurdles making research challenging for space robotic applications using RL and machine learning, particularly due to insufficient resources for traditional robotics simulators like CoppeliaSim. Our solution to this is an open source modular platform called Reinforcement Learning for Simulation based Training of Robots, or RL STaR, that helps to simplify and accelerate the application of RL to the space robotics research field. This paper introduces the RL STaR platform, and how researchers can use it through a demonstration.

## 1 INTRODUCTION

Reinforcement learning (RL) is a branch of machine learning, employs an iterative process of trial and error through direct interaction with the environment. It holds great promise to increase robotic autonomy for difficult real world tasks in both structured, and particularly unstructured environments. Unstructured environments have been an area of interest for researchers due to insufficient modeling and costly computational requirements which can hinder traditional approaches. Machine learning has seen an increase in popularity within recent years, with certain applications garnering significantly more attention, such as computer vision. RL by comparison has received less attention but has still made great progress, as shown by the number of publications for each subfield.

A quick search on arXiv, a popular free paper database for preprints commonly used in the machine learning field, shows this disparity[1]. By searching "Computer Vision" and "Reinforcement Learning" keywords, one can find 48,385 results for computer vision as opposed to only 7,697 results for reinforcement learning throughout the entire database. If narrowed down to just the past year, it shows 17,599 cases and 3,675 cases respectively. This could signify an increasing interest emerging in the RL branch of machine learning.

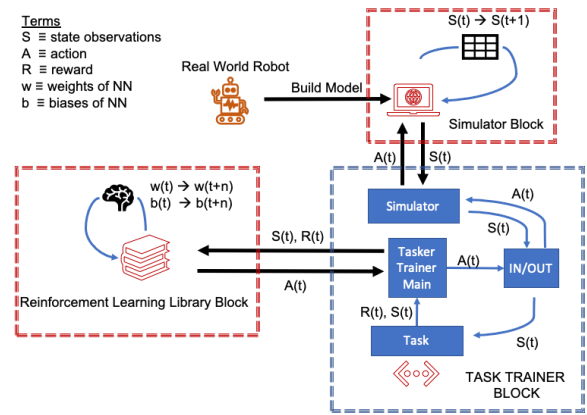

*Figure 1: High level description of the RL STaR platform showing the different blocks of the platform (denoted by the dotted boxes) and the transfer of information between them.*

Within reinforcement learning, there are several well-known applications on video games, outperforming human ability in both old and new games alike, including Go and Dota 2[2],[3]. While Go has a relatively simple action space, you put down one piece in an open area during your turn, it has a large number of possibilities, games lasting a couple hundred turns, creates over 10e170 possible state values for the board positions[4]. On the contrary, Dota 2 is focused on high-speed realtime teamwork in a 5v5 setting. Between the two, RL has been shown to excel in adversarial games that are both discrete, i.e. turn based, and continuous, as well as in both individual and team based settings. However,

RL applications outside video games are also numerous and robotics research is seeing rising interest recently, with applications for both high level functions such as decision making[5] and low level functions such as controlling a robotic hand[6].

There are two main ways to train RL for robotics. The first is to train directly on the robot. In this case, the learning algorithm is run directly on the robot, so the robot interacts with the environment physically. Any data gathered by its sensors or periphery sensors can be collected, aggregated together, and used by the robot to update its agent. An agent is a term used to describe a RL based decision making system. This approach has the benefit of training in a more realistic environment that can perhaps even be the actual environment you want the robot to function in. However, training using the onboard CPU of a robot can present issues due to lower available computation power as compared to a workstation or a cloud computing service. Additionally, this can also be dangerous for the hardware, i.e. a bipedal robot learning to walk and falling over can break. Thus, a second approach of training using a simulated environment, and then transfering the results to the actual robot has increased in popularity. This transfer is termed simulation to reality transfer, or sim2real for short. This method has the reverse benefits and disadvantages compared to training on an actual robot, namely being safer for the robot, faster for training, and giving us the ability to train in environments we might not have access to, while on the other hand suffering from inaccuracies and limitations of the simulation.

There are a number of platforms and simulators that have been used for 3D robot applications. That said, many of the simulators for these efforts are different from the ones traditionally used in the robotics community, such as CoppeliaSim and Gazebo. This makes it harder for robotics researchers to incorporate RL into their work, while also potentially increasing the gap between the real world and the simulation, termed the sim2real gap[7],[8]. This gap determines if a RL trained robot, which works in simulation, will also work in the real world, and how similar the actions between the two will be.

We created the Reinforcement Learning for Simulation Based Training of Robots Platform, or RL STaR for short in order to make RL more accessible for robotics researchers using CoppeliaSim. This platform combines the three main components needed for reinforcement learning in robotic applications: the RL algorithm, the simulator and a modular component that connects the two aforementioned parts with the task to be learned.

## 2 BACKGROUND

### 2.1 Reinforcement Learning

Reinforcement Learning (RL) is a branch of machine learning that differs from the others in that the agent generates its own training data. It does this by interacting with the environment, whether it is in the real world or in a simulation. RL teaches an agent how to solve a certain problem via optimizing via a reward function. The reward function is traditionally set before the training problem, and gives a reward to the agent based off of its transitions between states. States are the collection of data relevant for the agent to make an action and to describe its current condition. This state information, when given to the agent, is called an observation. This information can be either perfect, termed fully observable, or imperfect, i.e. noisy or only partially available, and thus termed partially observable.

The goal of the agent is usually to maximize the reward function, but depending on the algorithm, several other features or objectives can be incorporated as well, such as increasing entropy, or the spread of the probability over multiple actions for a given state. Such Features are popular in many modern algorithms, whether it be entropy or Kullback-Leibler (KL) divergence, also known as relative entropy[9]. These features often are meant to spur additional exploration in order to overcome local maximum in a search for either a higher local maximum or a global maximum. There have been some studies showing their possible effectiveness, often empirically[10].

### 2.2 Reinforcement Learning in Simulators

Simulators help us model the real world and physical interactions. This is important for robotics, as it allows us to test control applications and different robot configurations, such as their kinematics. It allows for faster prototyping and helps us create algorithms in a safer and efficient manner. There are several important components of simulators, the first is how it propagates everything forward in time. This is typically done by a physics engine, which estimates the force interactions and states of all the objects. Naturally, an ability to create the objects and the environment or world that you desire is also critical. Lastly, a GUI and visualization is also beneficial to observe how the system interacts and evaluate the effectiveness of what you are trying to achieve. Thus, you can create objects within the environment, such as a floor or a wall, along with a robot, such as a lunar rover, and test the interaction between the two.

Several efforts have gone into streamlining how RL environments are made, such as Gym, which was created by OpenAI in order to standardize the way RL problems are set up, making it easier to share code and compare results amongst researchers[11]. Every Gym environment has the same skeleton of basic functions and formatting, which includes a declaration of the dimensionality and scale of the actions and observations, amongst other things. It includes some structure which dictates how to reset and set up the problem.

## 2.3 Robots in Unstructured Environments

Certain applications are characterized by unstructured and unknown environments, such as space robotics, field robotics and disaster relief robotics. These areas have several factors that make them unstructured which can include: a limited degree of apriori known ground conditions, uncertain deployment/travel destination, uncertain task definition, and stochastic conditions, such as variable friction that might be hard to model. These cases often enable us to define the problem as a Markov Decision Process (MDP). MDPs are problems with some inherent uncertain or probabilistic transition between one state to the next, either in the form of multiple state possibilities given a certain action or multiple reward possibilities given a certain action. These must meet the markov property, which states that the transition probability at time t is independent of all the previous timesteps. In modern times, RL has been used to try to optimize solutions for many of these problems.

## 3 PRIOR AND RELATED WORKS

There have been many simulators developed for RL applications, and it is worth briefly discussing them.

Multi-Joint dynamics with Contact, or MuJuCo, is a physics engine that has been popularly used for reinforcement learning applications[12]. While popular, it has received criticism in the community for not being free, and as such, alternatives like Roboschool have popped up. MuJuCo claims to be faster and more accurate for certain types of interactions as compared to other engines such as Bullet, which could be a benefit of this physics engine and simulator[13]. It comes with several robot-like models, such as "humanoid", a 2-legged human like model, and "ant", a 4-legged model, that can be used for various RL tasks, such as teaching the model how to walk. MuJuCo's graphic user interface (GUI) is relatively easy to use and is traditional for the robotics community with xml format.

OpenAI's Roboschool was launched as a free alternative to MuJoCo[14]. This platform contained a default flat rectangular world with football field graphics placed on top. It includes several robot-like bodies similar in appearance to MuJoCo's, such as humanoid and ant, although with different characteristics such as weight. It couples the GUI with the Bullet physics engine, which is generally considered fast but not the most accurate, and custom code is available to train the agents to walk forward. Other tasks include chasing goals and walking under force disturbances (being pelted by cubes). This was recently discontinued, citing the success of PyBullet, a free alternative, as a reason[15]. While Roboschool contained a great GUI, it was not easy to change the environment, such as a lack of interactive interface for adding objects, which limited its applicability for the robotics community. PyBullet is also well built and uses the Bullet engine. It has a GUI that is relatively easy to use and familiar for the robotics community, however, has not traditionally been used. It uses OpenGL for rendering and can load URDF and SDF files, two file formats for modeling objects by code.

CoppeliaSim and Gazebo are two simulators commonly used in the robotics community[16]. Within CoppeliaSim, there are a number of options for physics simulators, including Bullet. One of the reasons that many machine learning researchers tended to overlook the more traditional simulators was because many machine learning researchers came from outside the robotics community. Another reason was due to overall speed. Recently, the speed concern was addressed in PyRep, which is a toolkit aimed at, among other things, increasing CoppeliaSim's speed for learning tasks[17]. This should make the platform more attractive for both robotics focused researchers and other machine learning researchers alike.

Pyrep is a recently released toolkit aimed at making CoppeliaSim more appealing for the typical machine learning researcher through speed improvements, with a new rendering engine and API improvements[17]. This recent upgrade illustrates the growing interest and understanding within the robotics community, as well as the robotics simulator's makers' acknowledgement about the gap that currently exists between machine learning and robotics.

Baselines is a library of RL algorithms created by OpenAI. It includes a number of modern algorithms, such as Actor Critic using Kronecker-Factored Trust Region (ACKTR) and Proximal Policy Optimization (PPO)[9], [14].

Some work has also been done on the Gazebo simulator front in order to encourage greater usage. The openai_ros package makes use of the OpenAI platform to train ROS based robots on the Gazebo simulator[18]. It does so by providing users with a Gazebo Environment class that enables all necessary connections between Gazebo and OpenAI Baselines for training, with the help of the ROS communication architecture. This added layer is universal for any project and shares training information through ROS topics. Two other modules, the Task Environment and the Robot Environment classes, are also part of this package and can help to build a robot training project from scratch.

Many researchers have shown an interest in RL and machine learning for robotics, as well as for space robotics. This includes a great number of robot types, such as walking robots, tensegrities, and rovers[19]–[22]. Applications have included learning to walk on variably sloped terrains, learning locomotion in rough terrains and using computer vision and machine learning for enhancing autonomy on planetary rovers. While path planning is a frequent subject in RL, it is often done by using set navigation modes instead of a motion control trained specifically with this goal in mind[23]. Nonetheless, motion control is a hot topic on its own for AI developers, with recent research on the attitude motion control of humanoid robots and solving strong coupling nonlinear problems[24],[25].

## 4 RL STaR PLATFORM

### 4.1 Platform Structure

The platform was created in a modular way such that it will easily evolve over time, be simpler to understand and open to collaborative work. This can be seen in how we broke down the structure into 3 mains blocks, one for the RL algorithms, one for the simulator, and one for the actual application, called the Task Trainer Block. This can also be seen with how we broke down the Task Trainer Block into subblocks.

The three main blocks of the platform (Fig. 1):

**1) RL Library Block:** this block is responsible for initiating the start of the simulation through a call to the run file, and specifies the algorithm to be used, along with the neural network (NN) architecture, some optimization parameters, and the number of training steps desired. This block contains a number of RL algorithms to select from when training. OpenAI's Baselines was chosen to ship with this platform[26]. This was due to its popular usage in the RL community, due to its professional appearance and the availability of a decent number of modern choices for RL algorithms. This library should be easily swappable with any other RL library that is compatible with the Gym environments though, such as Stable Baselines or Tensorflow RL Agents[27], [28].

**2) Simulator Block:** CoppeliaSim was used as the simulator due to its popularity within the robotics community, its relative ease of use and good graphics user interface. We wanted to ensure that the simulator would not limit the user in terms of setting up an interesting and practical environment to train the robot. Commands are received from the task trainer block to specify the next action of the robot and to step the simulation forward. The simulator block also passes back the state observation, information about the simulation and the robot state, to the task trainer block. Advanced features include detecting collisions, and adding various sensors such as cameras and lidar.

**3) Task Trainer Block:** an intermediary modular block that connects everything, defines the task, environment, and robot. This block is broken down into several sub blocks consisting of: IN/OUT, Simulator (API commands), Tasks, Main, and Constants (not shown in the diagram as it is only internal). This block is probably the one that researchers would spend most of their time customizing. It is also responsible for setting up some of the conditions of the simulator, such as specifying which file to load for the robot and world, and the action dimensions and observations to bring back from the simulator. The simulator sub block is responsible for communicating back and forth between this block and the simulator block using the python API.

The RL library block updates the neural networks (NNs) after processing each block of n timesteps, updating the weight and bias parameters. The simulation block propagates the simulation forward by timesteps. The task trainer block determines the reward based on the chosen task and the state values for a given timestep. Several of the task trainer sub blocks responsible for passing data are shown. Each sunblock (blue box) is contained within a separate file.

### 4.2 Task Trainer Block Modular Configuration

We broke the task trainer block into the multiple sub blocks, with each sub block contained in a file and having a unique function. This makes it faster to become familiar with this platform and to use it for machine learning robotics research. By separating the files, it also makes it simpler to find the section of code you are trying to change and minimizes the chances of breaking the code accidentally. Critically, this platform is generalizable and scalable, in order to be used by different researchers, for a variety of tasks, along with making it straightforward to upgrade the system in the future and for researchers to share their work.

### 4.3 Simulator, Setup, Robot Link

The CLOVER rover is a small four wheeled skid steering robot built by the Space Robotics Lab for multi-rover collaborative exploration research. We modeled it as an object in the simulator by simplifying the overall geometry into a number of simple shapes. This is done in order to save computation cost trying to keep the overall system dynamics the same. Several motors were then added as actuators and are attached to the wheels. The naming of each part is important, as this name is used as a unique identifier for communications between the Simulator Block (CopelliaSim) and the Task Trainer Block.

## 5 TRAINING TASK EXAMPLE

The main goal of this platform is to help researchers mix robotics, space, and machine learning/RL. This is primarily done by the tasks that the robots are trained to solve. These tasks can be broken down into sub modules and called upon during the training process based on which task you would like the robot to solve. This allows the researcher to change as few things as possible when defining the problem, and to share their tasks with other researchers. We will share an example of using the platform to train the CLOVER rover for a task.

### 5.1 Path Planning and Motion Control (PPMC)

The RL STaR platform is shipping with one task, dubbed the path planning and motion control task (PPMC) based on some prior work on a walking robot moving on flat terrain and a rover generalizing to hilly terrain[29], [30]. This task teaches the robot how to control its motors in order to move and turn to get to the user specified waypoints. It is learned in a model free manner, meaning that there is no learned or apriori produced assumption of how the system behaves. Rather, everything is learned through trial and error, accumulating experience and adjusting the actions accordingly. In other words, the agent learns how to operate the motors in order to maximize the reward, given for making progress towards the goal. Learning takes place when the reward function is suitable for the task it is we would like it to learn. In this example, we rewarded the agent for making progress to the current goal, and randomizing the location of the waypoint for each episode, ensuring that it does not just memorize one specific destination it needs to go to.

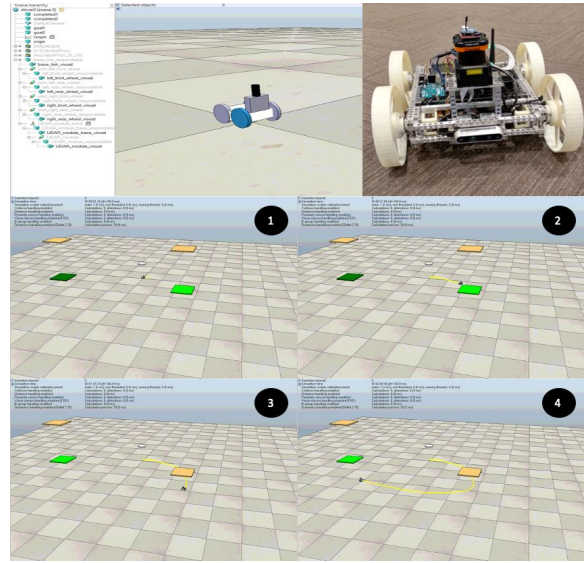

*Figure 2: A sequence of images showing a trained rover progressing through a first waypoint and proceeding to a second waypoint.*

### 5.2 Map and Rover Characterization

As a demonstration, we train the model CLOVER rover for the PPMC task on a flat terrain in a 10m x 10m grid. The rover is two wheel drive, with the front motor on each of the two sides controlling the speed of both the front and rear wheel. The agent action range is from [0,1]. This corresponds to a minimum velocity of 0 m/s and a maximum velocity of 0.1 m/s (when both motors were driven with an action of 1). An action multiplier was used to scale this output to either increase or decrease the maximum speed of the rover. We trained the agent with an action multiplier of 2, corresponding to a maximum velocity of 0.2 m/s. The actual motor-restricted maximum speed of the rover is 0.4 m/s, however, in practice for lunar applications, the rover is not expected to ever reach such high speeds. Note that we set the minimum speed to 0 m/s as there was no need for the rover to reverse during this task.

Contrasting some prior work, for this demonstration, we aimed to minimize the number of states we collect to use for the observations in the training of the RL agent[29], [30]. This information is placed into an array and fed to the two neural networks. Actor-critic type RL algorithms, such as PPO and ACKTR, have two neural networks, one that determines the next action given the current state, and a second one that determines how optimal certain states are given the predicted future rewards. This state array consists of: the x and y coordinate of the rover (relative to the origin), the velocity of the rover in the x and y direction relative to its base (forward and to the side), the angular yaw, as well as some information related to the current waypoint. This waypoint information held information about the angle between the rover and the waypoint, the distance to the waypoint, and the waypoint x and y coordinate. Thus in total there are 9 states, 5 states relative to the rover, and 4 relative to the current waypoint. All the states were normalized to a range [-1,1] between their expected [min, max] values.

### 5.3 Tuning Process

The tuning process is an important step in the modern reinforcement learning and machine learning methodology. Three important things needed to be tuned for this example: the RL algorithm, the neural network and the reward function.

*Table 1: Parameters and their values used for tuning the PPO algorithm for the demonstration*

| Parameter | Value | Parameter | Value |
|---|---|---|---|
| learning rate | 3e-4 | number of steps per update | 256 |
| discounting factor | 0.9 | advantage estimation discounting factor | 0.95 |
| entropy coefficient | 0.01 | value function coefficient | 0.5 |
| number of training epochs per update | 4 | number of minibatches per update | 4 |
| clipping range | 0.2 | max grad norm | 0.5 |

For this demonstration, PPO from Baselines was used, however, it was also tested with ACKTR with different parameters. The Baselines RL Libraries code is well documented with the tunable parameters for each algorithm, giving acronyms and the meaning of each variable[26]. In Tab. 1 we display the values we used for the training of the agent with PPO.

The neural network architecture consisted of a fully connected diamond shaped neural network, with 5 layers sized: 64x128x164x128x64, tanh activation function, and without layer normalization.

### 5.4 Reward Function

The reward function, R(G,t), is calculated at each timestep with respect to the goal and has 3 main components: the primary reward, P, meant to directly encourage the task to be learned; beneficial rewards, B, meant to encourage good habits; and detrimental penalties, D, meant to discourage bad habits[30]. In the case of this demonstration, the primary reward is a function proportional to the progress made towards the current waypoint, either positive if the agent moved closer, or negative if the agent moved further. There are no beneficial rewards in this case. There are two detrimental penalties, one constant value given at each timestep to discourage slow movement and a second as a function proportional to the yaw rate (where $\theta$ is yaw), to discourage superfluous rotations. The episode ends when either two waypoints have been reached or the time limit has passed. If a goal is reached, we give a bonus reward to the rover by assuming that it made progress of 0.5 m in the last timestep, substituting $X$ for 0.5 in Eq. 1.

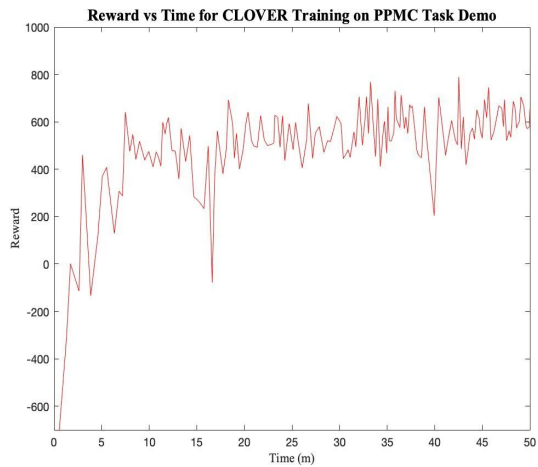

*Figure 3: Reward curve obtained for the demonstration. Each point represents the average of 5 batches, each with 256 timesteps. The X axis represents real-time elapsed since training began.*

$$R(G, t) = P + B - D = C_{veloc}X - C_{alive} - C_{turn}\frac{d\theta}{dt} \quad (1)$$

Where constants $C_{veloc}$, $C_{alive}$ and $C_{turn}$ are equal to 50, 0.5 and 1, respectively.

As shown in the reward curve in Fig. 3, there are three or so distinct phases of the learning process. The first is the primary learning, where the agent learns to accomplish the main goal of the task, and is discernible by a steep gradient in the reward curve. The second phase is the optimization phase, where the agent already achieves the primary objective, but can still optimize to maximize the beneficial reward and minimize the detrimental penalties. This is discernible by a shallow but still significant slope in the reward curve. The last phase is when the agent has plateaued.

## 6 CONCLUSION

This paper introduces a platform for applying RL to robotics and space robotics in a manner accessible to many more traditional researchers. This is done through a combination of the CoppeliaSim, the Baselines RL libraries and a task training block. The objective is to make it simpler for robotics researchers who might not otherwise apply RL to their problems to do spo, and to then share their work with other such researchers. This platform was made in a modular way to allow for sharing and collaboration, to be straightforward to use and simple to build upon.

One additional modular block that would be of high interest to robotics researchers is a ROS block. This could be in place of the IN/OUT block, and allow ROS messages to control the robot, as opposed to direct messages from the RL algorithm, which might be a desirable step before conducting sim2real transfer. As RL STaR was released with just one robot and one task, the addition of various robots and tasks as more people use the platform could help encourage creativity and innovation in the robotics RL community. This will include high-level decision making tasks, such as searching for resources and exploring an area. More robots can be added and shared with each other, such as high speed rovers for lunar explorations[31]. Additionally, more complicated environments can be added for and by the community, such as hilly or obstacle rich environments. Such environments will also require the addition of sensors such as LIDAR or cameras. This can stimulate new solutions to synergistically combine computer vision and RL for robotics. Lastly, this implementation of RL star uses the original CoppeliaSim renderer, and so perhaps by upgrading to the Pyrep version, faster simulation can be enabled, which could be advantageous for some applications.